\documentclass[11pt]{article}

\usepackage{acl}

\usepackage{times}
\usepackage{latexsym}

\usepackage[T1]{fontenc}
\usepackage[utf8]{inputenc}

\usepackage{microtype}

\usepackage{inconsolata}

\usepackage{graphicx}

\title{{MasalBench: A Benchmark for Contextual and Cross-Cultural Understanding of Persian Proverbs in LLMs}}

\author{
  Ghazal Kalhor\textsuperscript{1} \quad
  Behnam Bahrak\textsuperscript{2}
  \\
  \textsuperscript{1}EPFL \quad
  \textsuperscript{2}Khatam University
  \\
  \small{
    \textbf{Correspondence:}
    \href{mailto:ghazal.kalhor@epfl.ch}{ghazal.kalhor@epfl.ch},
    \href{mailto:b.bahrak@teias.institute}{b.bahrak@teias.institute}
  }
}

\usepackage{booktabs}
\usepackage{xltabular}
\usepackage{array}
\usepackage{caption}

\begin{document}
\maketitle

\begin{abstract}
In recent years, multilingual Large Language Models (LLMs) have become an inseparable part of daily life, making it crucial for them to master the rules of conversational language in order to communicate effectively with users. While previous work has evaluated LLMs' understanding of figurative language in high-resource languages, their performance in low-resource languages remains underexplored. In this paper, we introduce \textsc{MasalBench}, a comprehensive benchmark for assessing LLMs' contextual and cross-cultural understanding of Persian proverbs, which are a key component of conversation in this low-resource language. We evaluate eight state-of-the-art LLMs on \textsc{MasalBench} and find that they perform well in identifying Persian proverbs in context, achieving accuracies above 0.90. However, their performance drops considerably when tasked with identifying equivalent English proverbs, with the best model achieving 0.79 accuracy. Our findings highlight the limitations of current LLMs in cultural knowledge and analogical reasoning, and they provide a framework for assessing cross-cultural understanding in other low-resource languages. \textsc{MasalBench} is available at \url{https://github.com/kalhorghazal/MasalBench}.
\end{abstract}

\begin{figure}[!t]
  \centering
  \includegraphics[width=0.87\columnwidth]{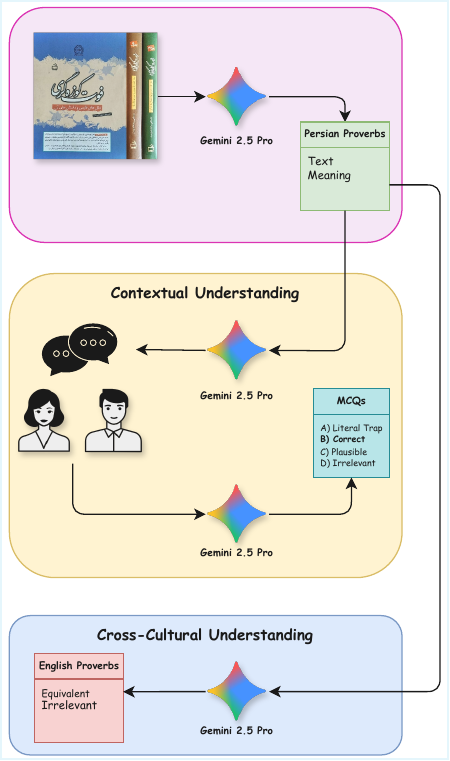}
  \caption{Pipeline illustrating the construction of the benchmark for assessing LLMs' contextual and cross-cultural understanding of Persian proverbs.}
  \label{fig:pipeline}
\end{figure}

\section{Introduction}

Large Language Models (LLMs) are increasingly used in daily life, from offering product advice to answering general questions and supporting business tasks. To communicate effectively, they must understand users' conversational language. Proverbs, traditional sayings that convey common-sense truths metaphorically, are an important aspect of natural conversation. Multilingual LLMs perform well on proverbs in high-resource languages, particularly English, thanks to abundant training data and their inclusion in benchmarks \citep{kim2025memorization, fu2025chengyu, almeida2025broverbs, liu2024multilingual}. In contrast, the scarcity of such data in low-resource languages may limit LLM proficiency with their proverbs, and only a few studies develop benchmarks to investigate this capability, typically focusing on a small and limited set of low-resource languages \citep{azime2025proverbeval, thapa2025probing, donthi2025improving, nguyen2025don, das2026meaning}.

Persian is one of the low-resource languages for multilingual LLMs, yet it is exceptionally rich in the use of proverbs, both historically and in contemporary contexts. Iranian speakers naturally employ proverbs frequently in daily conversations and even in professional settings, often without being consciously aware of it \cite{hadissi2010socio}. However, only a few studies examine LLMs' ability to understand Persian proverbs. For example, \citet{khoshtab2024comparative} propose 316 binary-choice questions in dialogue format, where each question pairs the correct interpretation of a proverb with a single incorrect option generated by GPT-4o. In 60 of these questions, the incorrect option corresponds to the literal meaning of the proverb, making the evaluation of contextual and figurative understanding more challenging. Similarly, \citet{farsi2025melac} design 370 multiple-choice questions that ask for the meaning of Persian proverbs, where the distractors are meanings of randomly selected proverbs.

To address this gap, we introduce \textsc{MasalBench}, a comprehensive benchmark for evaluating LLMs' understanding of Persian proverbs at both contextual and cross-cultural levels. At the contextual level, we propose a set of 1,000 multiple-choice questions in dialogue form, where each of the three distractor options is carefully designed to serve a specific purpose: a literal trap, a plausible but incorrect meaning, and an irrelevant meaning. At the cross-cultural level, we propose 700 binary-choice questions to assess whether LLMs can identify the English equivalent of a given Persian proverb, with the distractor option crafted to share similar wording, tone, or imagery but convey a different meaning. While previous work explore proverb translation \citep{donthi2025improving, rezaeimanesh2025large}, to the best of our knowledge, no prior study has captured the cross-cultural dimension through systematic equivalence pairing between languages. Finally, we evaluate \textsc{MasalBench} on eight prominent multilingual LLMs.

Building on these contributions, this study investigates three research questions: \textbf{RQ1:} How well do multilingual LLMs understand Persian proverbs in context? \textbf{RQ2:} What kinds of mistakes do they make when failing to do so, and how often do these correspond to literal, plausible, or irrelevant errors? \textbf{RQ3:} To what extent can LLMs identify the correct English equivalent of a given Persian proverb?

Our results indicate that multilingual LLMs perform strongly in understanding Persian proverbs in context, with model size and instruction tuning being key contributors. When errors occur, they predominantly involve plausible distractors, reflecting the models' tendency to select semantically coherent interpretations, a pattern more pronounced in reasoning-focused models. In contrast, identifying the correct English equivalent of a Persian proverb proves more challenging, and although instruction tuning provides some benefit, its impact is limited due to the need for cultural knowledge and analogical reasoning.

\section{Benchmark Construction}

To construct our benchmark, we collect Persian proverbs and their meanings from \textit{Foote Koozegari: Persian Proverbs and Their Stories} \citep{rahmandoost2011foote}, a comprehensive reference that documents the etymologies, narratives, and interpretations of over 4,000 Persian proverbs. For this task, we use Gemini 2.5 Pro \citep{comanici2025gemini} with the PDF versions of the book (both volumes), as the texts are provided in image format rather than as digital text.\footnote{Whenever we use Gemini 2.5 Pro for automated content generation, we set the temperature to 1 and top-p to 0.95 to ensure a balance of creativity and coherence.} We choose Gemini because it has strong performance on Persian text recognition and understanding,\footnote{\url{https://mcinext-mizan-llm-leaderboard.hf.space/}} making it well suited for processing scanned Persian texts. To ensure consistency during benchmark construction, we use a single model and manually review the extracted content to correct potential typos. We then select 1,000 proverbs that are widely used and well known in Iranian culture, based on the authors' cultural familiarity and their frequency of occurrence in everyday language, literature, and media. All manual reviews and verifications are carried out by the authors, who are native Persian speakers familiar with the language and its cultural context. The overall pipeline for constructing the benchmark is illustrated in Figure~\ref{fig:pipeline}.

\subsection{Contextual Understanding}

To prepare the dataset for this task, we follow two steps. First, for 100 of our proverbs, we manually design short, natural, and conversational dialogues in two parts between two people. In each dialogue, the second person uses the given proverb in response to the first person. We also provide an explanation of the second person's purpose for using the proverb. We then automate this step for the remaining proverbs using Gemini 2.5 Pro, by providing the proverbs and their meanings. Second, for 40 dialogues, we manually design three distractor interpretations of the proverb: (1) a literal interpretation, (2) a logical but incorrect interpretation, and (3) an unrelated interpretation. We then automate this step for the remaining dialogues using Gemini, by providing the dialogue, the question (asking for the second person's purpose for using the proverb), and the correct purpose. All automatically generated dialogues and distractor options are manually verified by the authors for correctness and naturalness. These steps result in a dataset of 1,000 multiple-choice questions for assessing LLMs' understanding of proverbs in context. 

\subsection{Cross-Cultural Understanding}

To construct the dataset for this task, we provide Gemini 2.5 Pro with each Persian proverb and its meaning, and instruct it to generate two well-known and recognized English proverbs: (1) an equivalent proverb and (2) an irrelevant proverb. The equivalent proverb is most similar in meaning and usage to the given Persian proverb, while the irrelevant proverb has no real connection in meaning or usage to the Persian proverb, but may appear similar in form (for example, in terms of words, tone, or imagery), and therefore serves as a convincing distractor. If no clear or culturally appropriate English equivalent exists for a given Persian proverb, we exclude that proverb from this task and instead sample additional proverbs from the source collection. We then manually verify the results to ensure that the generated proverbs are authentic English expressions and that the equivalents accurately match the Persian proverbs in meaning. Additionally, for each proverb, we add a question asking which English proverb is equivalent to it. Through this process, we compile a dataset of 700 binary-choice questions that assesses LLMs' cross-cultural understanding of proverbs.

\section{Experiments}

\subsection{Language Model for Evaluation}

We evaluate several well-known multilingual LLMs of varying sizes with proficiency in Persian, including Llama 4 Scout \citep{meta2025llama4scout}, Llama 3.3 70B Instruct \citep{meta2024llama3}, Qwen 2.5 72B Instruct \citep{qwen2025technical}, Qwen QwQ 32B \citep{qwen2025qwq}, DeepSeek V3.1 \citep{deepseek2025v31}, DeepSeek R1 \citep{deepseek2025r1}, GPT-4.1 mini \citep{openai2025gpt41}, and GPT-4o mini \citep{openai2024gpt4omini}, on our benchmark.

\subsection{Prompting Setup}

To design the prompts for both tasks, we randomize the order of the options to avoid positional bias  \citep{pezeshkpour2024large}. We employ zero-shot prompting, and each prompt runs independently, with the LLMs not performing any web searches to answer them. We set the temperature to 0 to ensure deterministic responses without added creativity, and top-p to 1 so that all possibilities are considered. The maximum number of tokens is limited to 5, as the LLMs are expected to return only the letter corresponding to the correct option. Appendix \ref{sec:examplePrompts} provides example prompts used to construct \textsc{MasalBench} and evaluate the performance of LLMs on this benchmark.

\subsection{Evaluation Metrics}
We use accuracy as the primary evaluation metric. Since both tasks are single-label classification problems in which each example has exactly one gold label and each model prediction corresponds to a single label, micro-F1 is mathematically equivalent to accuracy. Therefore, we report accuracy only.

\section{Main Results}

\subsection{Contextual Level}

Table~\ref{tab:contextAcc} presents the accuracy of different LLMs on contextual understanding. As shown, all models perform highly on this task, achieving accuracy above 0.90, which indicates that contextual understanding of Persian proverbs is a relatively easy task for these state-of-the-art LLMs and that they are already robust at metaphorical reasoning in non-English contexts. DeepSeek V3.1 turns out to be the best-performing model with an accuracy score of 0.943, which could be explained by the fact that it has the largest size (671B parameters) among the models. However, size is not the only determining factor of LLM performance on this task, as the lowest accuracy score is assigned to Qwen QwQ 32B, even though it has a larger parameter size compared to the GPT models. We also observe that instruction-tuned models perform better than their reasoning-focused counterparts within the same families, which suggests that instruction tuning plays a significant role in improving cultural contextualization.

\begin{table}[h!]
\centering
\small
\begin{tabular}{ll}
\toprule
\textbf{Model} & \textbf{Accuracy} \\
\midrule
Llama 4 Scout & 0.906 \\
Llama 3.3 70B Instruct & 0.927 \\
Qwen 2.5 72B Instruct & 0.911 \\
Qwen QwQ 32B & 0.903 \\
DeepSeek V3.1 & \textbf{0.943} \\
DeepSeek R1 & 0.927 \\
GPT-4.1 mini & 0.920 \\
GPT-4o mini & 0.920 \\
\bottomrule
\end{tabular}
\caption{Accuracy values for contextual understanding across LLMs.}
\label{tab:contextAcc}
\end{table}

\begin{figure}[h] % [t] for top, [h] for here, [b] for bottom
    \centering
    \includegraphics[width=0.8\columnwidth]{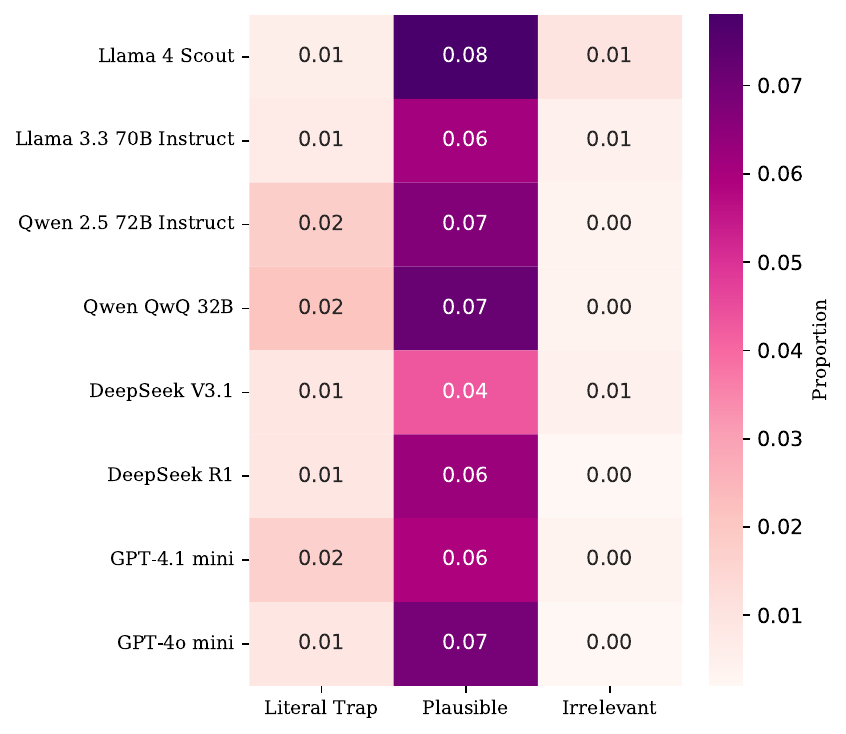} % replace with your PDF file
    \caption{Heatmap of the proportion of distractor option selections for contextual understanding across LLMs.}
    \label{fig:heatmap}
\end{figure}

To assess the types of mistakes LLMs make when failing to select the correct meaning of a Persian proverb in context, we analyze the proportion of responses corresponding to each distractor option. As shown in Figure~\ref{fig:heatmap}, the majority of errors across all models fall into plausible distractors. This suggests that when LLMs are uncertain about a proverb's meaning, they still attempt to choose a semantically coherent interpretation relevant to the dialogue. We also observe that within the same model family, reasoning-focused models are more likely to fall for plausible distractors compared to instruction-tuned models. In contrast, literal and irrelevant distractors occur at much lower rates (ranging from 0.02 to 0.00), indicating that these advanced LLMs rarely make such simplistic mistakes.

\subsection{Cross-Cultural Level}

Table~\ref{tab:culturalAcc} presents the accuracy of different LLMs on cross-cultural understanding. As shown, LLMs perform worse on this task compared to contextual understanding. This may be partly explained by the fact that cross-cultural understanding requires cultural knowledge beyond linguistic decoding and also demands analogical reasoning, since equivalent proverbs often employ different imagery or wording. The best-performing model is DeepSeek R1, with an accuracy score of 0.793, while the lowest-performing model is Llama 4 Scout (0.647). We also observe that instruction tuning continues to provide advantages in this task, although to a lesser extent than in contextual understanding, as success here depends more on training exposure to cultural data than on improved instruction following. Overall, these results suggest that LLMs are relatively strong at intra-language contextual reasoning but weaker at cross-cultural abstraction.

\begin{table}[h!]
\centering
\small
\begin{tabular}{ll}
\toprule
\textbf{Model} & \textbf{Accuracy} \\
\midrule
Llama 4 Scout & 0.647 \\
Llama 3.3 70B Instruct & 0.710 \\
Qwen 2.5 72B Instruct & 0.691 \\
Qwen QwQ 32B & 0.691 \\
DeepSeek V3.1 & 0.703 \\
DeepSeek R1 & \textbf{0.793} \\
GPT-4.1 mini & 0.703 \\
GPT-4o mini & 0.651 \\
\bottomrule
\end{tabular}
\caption{Accuracy values for cross-cultural understanding across LLMs.}
\label{tab:culturalAcc}
\end{table}

\section{Conclusion}
This work introduces \textsc{MasalBench}, a benchmark designed to probe the contextual and cross-cultural understanding of Persian proverbs by LLMs. The dataset includes 1,000 multiple-choice questions targeting speaker intent in context and 700 binary-choice items on matching Persian proverbs to their English equivalents. Our evaluation across eight widely used multilingual LLMs shows that they generally perform well in capturing the intended meaning of Persian proverbs in context, with larger model size and instruction tuning playing a crucial role in this strength. When errors do arise, they are typically due to plausible but incorrect alternatives, highlighting a consistent preference for semantically coherent interpretations, a tendency especially evident in reasoning-oriented models. By contrast, transferring this knowledge across cultures proves substantially harder: aligning Persian proverbs with their English counterparts remains a notable challenge, as it calls for cultural grounding and analogical reasoning beyond what instruction tuning alone can provide. Taken together, \textsc{MasalBench} and our findings shed light on current limitations and offer a foundation for advancing research on cross-lingual and culturally informed language understanding. %A discussion of limitations and future research directions is provided in Appendix \ref{sec:limits}.

\section*{Limitations}%\label{sec:limits}

Our study has several limitations that could be addressed in future work. First, we rely solely on a single LLM (Gemini 2.5 Pro) for constructing our benchmark, even though other prominent LLMs with proficiency in Persian exist. Future research could explore alternative LLMs and compare the quality of their generated questions with ours to identify the most suitable model for this task. Second, equivalent English proverbs for Persian proverbs are relatively rare, making our cross-cultural understanding task inherently more challenging for LLMs than the core task itself. Future work could investigate alternative methods for evaluating LLMs' cross-cultural understanding of Persian proverbs.

%\section*{Acknowledgments}

\bibliography{custom}

\appendix

\section{Example Prompts}\label{sec:examplePrompts}
\subsection{Prompts for Benchmark Construction}
\begin{figure}[t]
  \centering
  \includegraphics[width=\columnwidth]{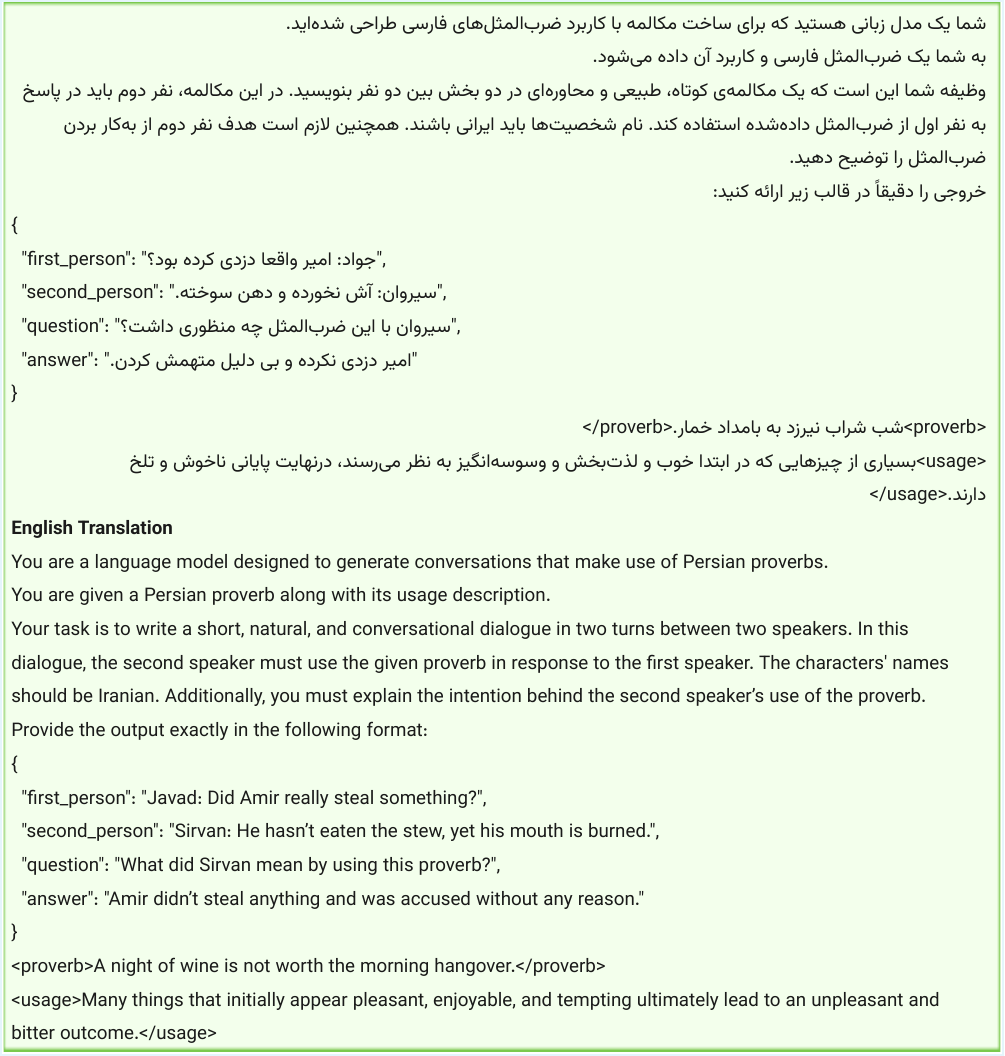}
  \caption{Example prompt for generating dialogues for Persian proverbs, along with its English translation.}
  \label{fig:dialoguePrompt}
\end{figure}

\begin{figure}[t]
  \centering
  \includegraphics[width=\columnwidth]{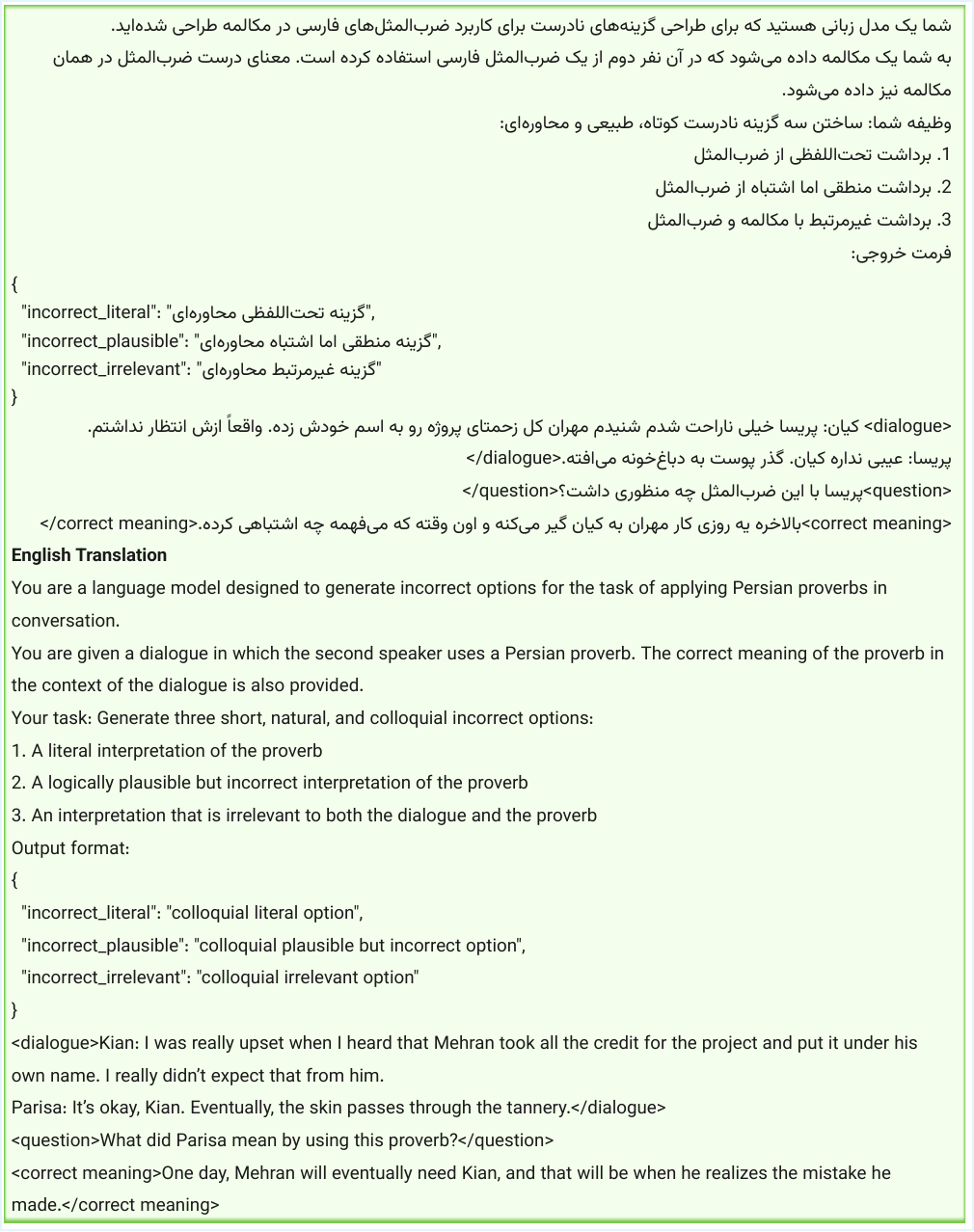}
  \caption{Example prompt for generating distractor options in multiple-choice questions for contextual understanding, along with its English translation.}
  \label{fig:wrongOptionPrompt}
\end{figure}

\begin{figure}[t]
  \centering
  \includegraphics[width=\columnwidth]{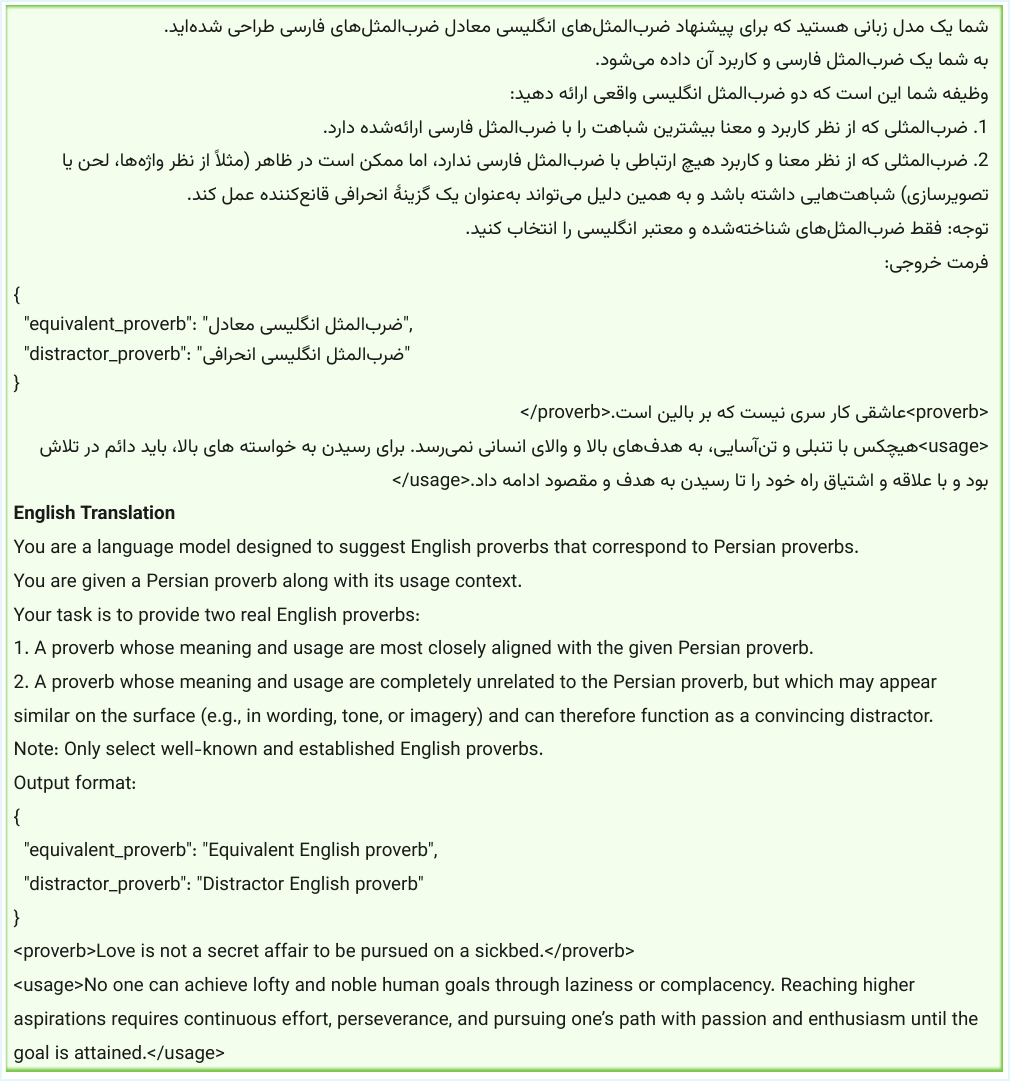}
  \caption{Example prompt for generating English equivalents of Persian proverbs for cross-cultural understanding, along with its English translation.}
  \label{fig:equivalentPrompt}
\end{figure}

Example prompts used in constructing our benchmark are provided in Figures~\ref{fig:dialoguePrompt}--\ref{fig:equivalentPrompt}, illustrating (1) dialogue generation with Persian proverbs, (2) distractor option generation for multiple-choice questions, and (3) equivalent English proverb generation.

\subsection{Prompts for Evaluation}
\begin{figure}[t]
  \centering
  \includegraphics[width=\columnwidth]{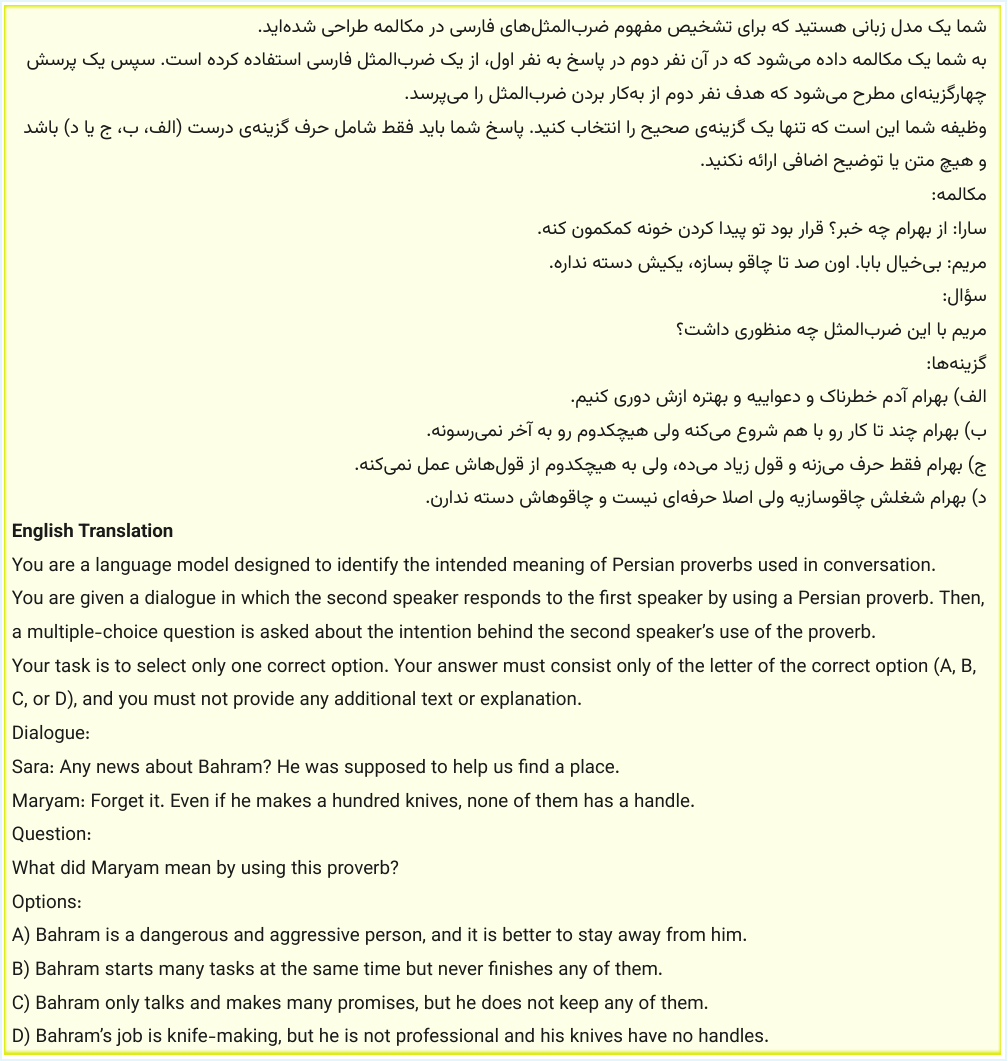}
  \caption{Example prompt for evaluating LLMs' contextual understanding of Persian proverbs, along with its English translation.}
  \label{fig:multipleChoiceQ}
\end{figure}

\begin{figure}[t]
  \centering
  \includegraphics[width=\columnwidth]{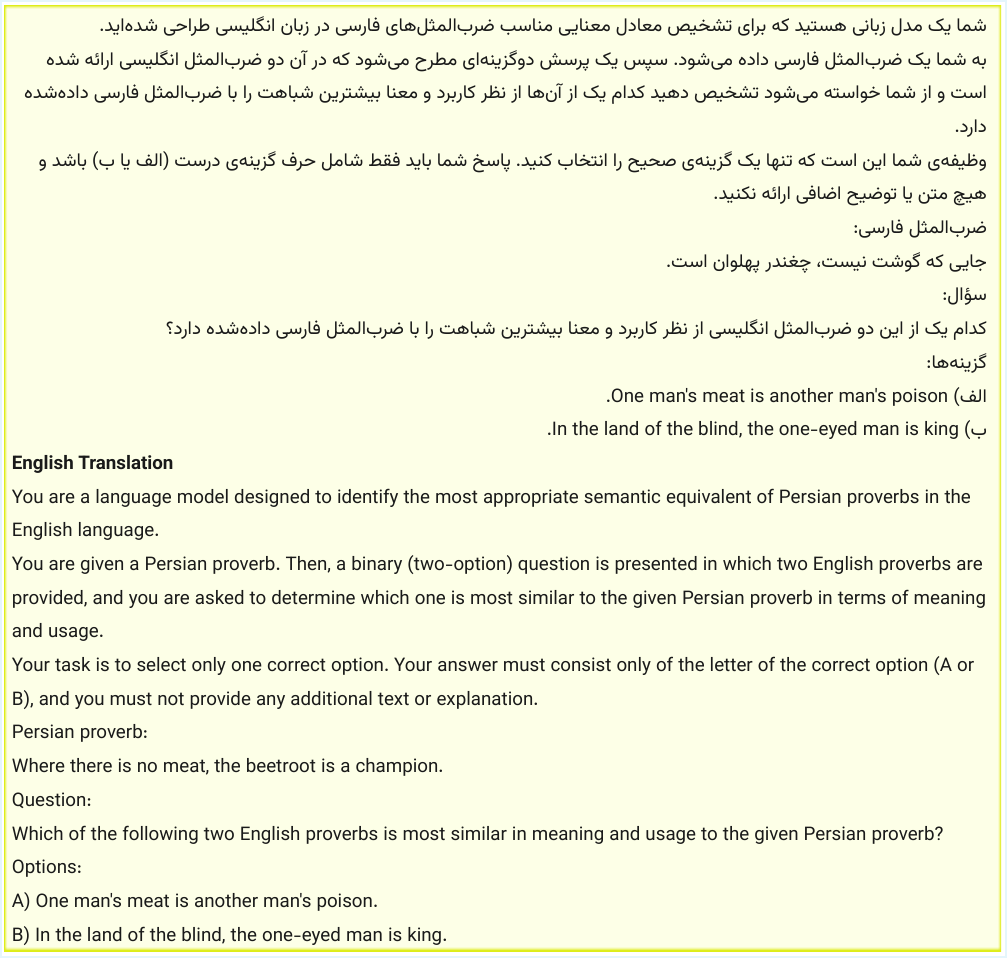}
  \caption{Example prompt for evaluating LLMs' cross-cultural understanding of Persian proverbs, along with its English translation.}
  \label{fig:binaryChoiceQ}
\end{figure}

Example prompts used to evaluate our benchmark on LLMs are provided in Figures~\ref{fig:multipleChoiceQ} and \ref{fig:binaryChoiceQ}, illustrating: (1) a multiple-choice question for contextual understanding, and (2) a binary-choice question for cross-cultural understanding.

\end{document}